\documentclass[sn-basic,iicol,NameDate]{sn-jnl}% Basic Springer Nature Reference Style/Chemistry Reference Style
\usepackage[nolist, nohyperlinks]{acronym}
\usepackage{algorithm}%
\usepackage{algorithmicx}%
\usepackage{algpseudocode}%
\usepackage{amsmath,amssymb,amsfonts}%
\usepackage{amsthm}%
\usepackage[title]{appendix}%
\usepackage{booktabs}%
\usepackage{catchfile}
\usepackage{graphicx}%
\usepackage{listings}%
\usepackage{makecell}
\usepackage{manyfoot}%
\usepackage{mathrsfs}%
\usepackage{multirow}%
\usepackage{natbib}
\usepackage{rotating}
\usepackage{subcaption}
\usepackage{textcomp}%
\usepackage{todonotes}
\usepackage{xcolor}%

\begin{acronym}
	\acro{AIOP}[AIO-P]{all-in-one predictors}
	\acro{BO}{bayesian optimization}
	\acro{CNN}{convolutional neural network}
	\acro{DARTS}{differentiable architecture search}
	\acro{ENAS}{efficient neural architecture search}
	\acro{FLOP}{floating-point operation}
	\acro{GNN}{graph neural network}
	\acro{GPU}{graphics processing unit}
	\acro{IoU}{intersection over union}
	\acro{LSTM}{long short-term memory}
	\acro{MAC}{multiply-accumulate}
	\acro{MDP}{Markov decision process}
	\acro{MNIST}{modified national institutes of standard and technology database}
	\acro{MORL}{multi-objective reinforcement learning}
	\acro{MTNAS}{multitask convolutional neural architecture search}
	\acro{NAS}{neural architecture search}
	\acro{NCP}{network coding propagation}
	\acro{PEB}{partial episode bootstrapping}
	\acro{PTB}{Penn treebank}
	\acro{RNN}{recurrent neural network}
	\acro{SSIM}{structural similarity index measure}
	\acro{SVHN}{street view house numbers}
\end{acronym}

\newcommand{\gciiacknowledgement}[0]{This research received funding from the Flemish Government (AI Research Program).}

\newcommand{\acknowledgement}[0]{\gciiacknowledgement\\\fwoacknowledgement}
\theoremstyle{thmstyleone}%
\theoremstyle{thmstyletwo}%

\theoremstyle{thmstylethree}%

\begin{document}

\title[Task Adaptation of Reinforcement Learning-Based NAS Agents Through Transfer Learning]{Task Adaptation of Reinforcement Learning-Based NAS Agents Through Transfer Learning}


\author*[1]{\fnm{Amber} \sur{Cassimon}}\email{amber.cassimon@uantwerpen.be}
\author[1]{\fnm{Siegfried} \sur{Mercelis}}\email{siegfried.mercelis@uantwerpen.be}
\author[1]{\fnm{Kevin} \sur{Mets}}\email{kevin.mets@uantwerpen.be}

\affil*[1]{\orgdiv{IDLab - Faculty of Applied Engineering}, \orgname{University of Antwerp - imec}, \orgaddress{\street{Sint-Pietersvliet 7}, \city{Antwerp}, \postcode{2000}, \state{Antwerp}, \country{Belgium}}}


\abstract{
    Recently, a novel paradigm has been proposed for reinforcement learning based \ac{NAS} agents, that revolves around the incremental improvement of a given architecture.
	We assess the abilities of such reinforcement learning agents to transfer between different tasks.
	We perform our evaluation using the Trans-NASBench-101 benchmark, and consider the efficacy of the transferred agents, as well as how quickly they can be trained.
	We find that pretraining an agent on one task benefits the performance of the agent in another task in all but 1 task when considering final performance.
	We also show that the training procedure for an agent can be shortened significantly by pretraining it on another task.
	Our results indicate that these effects occur regardless of the source or target task, although they are more pronounced for some tasks than for others.
	Our results show that transfer learning can be an effective tool in mitigating the computational cost of the initial training procedure for reinforcement learning-based \ac{NAS} agents.
}




\maketitle

\section{Introduction}
	Deep learning has made remarkable progress in recent decades across a wide range of domains including protein folding~\cite{Jumper_2021_Highly}, computer vision~\cite{He_2016_Deep,Tan_2019_Efficientnet}, language generation~\cite{Devlin_2019_BERT}, etc.
	A major driver of these improvements has been improvements in neural network architectures such as residual blocks~\cite{He_2016_Deep}, uniform scaling~\cite{Tan_2019_Efficientnet}, attention mechanisms~\cite{Vaswani_2017_Attention}, etc.
	As neural network architectures have gotten more and more complex, they require increasing amounts of researcher's and engineer's time to build and validate.
	Increasingly complex architectures often also come with increased computational costs.
	All of this makes the design of novel neural network architectures a complex, and time- and resource-consuming task.
	With this in mind, researchers have proposed to automate the neural network architecture design process in a research field called \acl{NAS}.
	In \ac{NAS} various techniques such as evolutionary algorithms~\cite{Elsken_2019_Efficient}, continuous relaxation~\cite{Liu_2019_Darts}, reinforcement learning~\cite{Pham_2018_Efficient}, etc.\ have been used to generate new neural network architectures.
	Some works only consider task performance~\cite{Liu_2019_Darts,Pham_2018_Efficient}, while others also consider parameters related to the required computational resources to run the neural network~\cite{Elsken_2019_Efficient,Tan_2019_MnasNet}.
	Much of the existing \ac{NAS} literature has focused on methods that approach a single task at a time.
	From a researcher's perspective, this a sensible approach, but from the perspective of a user of \iac{NAS} system, the costs of applying such systems to new tasks can make them unattractive.
	With this in mind, some research has been done into how \ac{NAS} systems can be designed that can easily design neural networks for multiple tasks~\cite{Ding_2022_Learning,Huang_2022_Arch,Mills_2023_AIOP,He_2024_Robustifying,Zhou_2024_Toward}.
	In this publication, we aim to contribute to this domain of task-transferable \ac{NAS} systems by enhancing the reusability of existing reinforcement learning-based \ac{NAS} agents across tasks.
	Specifically, we build on the work of \citet{Cassimon_2024_Scalable} with the aim of reducing the up-front of cost of training their \ac{NAS} agent.
	We take their reinforcement learning agent, and assess its merits and demerits in a transfer learning scenario, where an agent is first trained on one task, and then adapted for another task using transfer learning.
	We perform our assessment using the Trans-NASBench-101 benchmark~\cite{Duan_2021_TransNAS}.
	Specifically, we select a set of four different computer vision tasks and use these tasks to assess the transferability of our reinforcement learning agent.
	Within this scope, we consider three main research questions:

	\begin{enumerate}
		\item Does pretraining \iac{NAS} agent on one task benefit its training for another task?
		\item How well do agents with different transfer regimes perform on their target tasks, particularly with respect to zero-shot transfer, a fine-tuning regime and a full retraining regime?
		\item How much additional training is required to transfer a pretrained \ac{NAS} agent to a new task, and how does this relate to training an agent from scratch on this task?
	\end{enumerate}

	In Section~\ref{sec:sota} we provide a brief overview of recent (multi-task) \ac{NAS} methods and reinforcement learning transfer learning techniques.
	We follow this up in Section~\ref{sec:methods} by explaining our methodology in more detail.
	Next, details regarding our experimental setup along with an analysis of the obtained results are given in Section~\ref{sec:experiments}.
	Finally, we discuss our conclusions as well as some of the limitations of our work and possible directions for future work in Section~\ref{sec:discussion}.

\section{Related Work}
\label{sec:sota}

	\subsection{Reinforcement Learning-Based \ac{NAS}}
		\citet{Baker_2017_Designing} performed some of the first work in the field of reinforcement learning-based \ac{NAS}, with their work on MetaQNN.
		They make use of a tabular Q-learning-based methodology to design \acp{CNN}.
		Their agent operates in a chain-structured macro search space, targeting the CIFAR-10, \ac{SVHN} and \ac{MNIST} datasets.
		\citet{Baker_2017_Designing} achieve performance comparable to that of state-of-the-art neural networks at the time.

		Another pioneering work in the field of \ac{NAS} is that of \citet{Zoph_2017_Neural}.
		Similar to \citeauthor{Baker_2017_Designing}, they also make use of Q-learning to design neural networks.
		The approach used by \citet{Zoph_2017_Neural} differs in several key aspects though.
		The agent in~\cite{Zoph_2017_Neural} operates in both a macro and micro search space.
		They make use of \iac{LSTM}-based agent, rather than a Q-table and consider the CIFAR-10 and \ac{PTB} problems.

		\citet{Pham_2018_Efficient} improved on the work of \citeauthor{Zoph_2017_Neural} by introducing the concept of weight sharing.
		Using weight sharing, \citeauthor{Pham_2018_Efficient} were able to significantly cut down the computational resources required to find architectures with strong performance.
		Similar to \citeauthor{Zoph_2017_Neural}, they consider the CIFAR-10 and \ac{PTB} problems, and achieved state-of-the-art performance.

		Another work that uses reinforcement learning to build \ac{NAS} agents is that of \citet{Tan_2019_MnasNet}.
		They operate in a 2-objective setting, considering inference latency on mobile phones alongside CIFAR-10~\cite{Krizhevsky_2009_Learning} and Stanford Dogs~\cite{Khosla_2011_Novel} image classification accuracy.
		Since latency measurements on embedded devices are complex to set up and execute, proxies such as \acp{FLOP} or \acp{MAC} are usually used.
		In their work, however, \citeauthor{Tan_2019_MnasNet} measured the inference latency of the trained neural networks directly on a set of mobile phones, resulting in a much more accurate estimation of the inference latency.
		\citeauthor{Tan_2019_MnasNet} use \iac{LSTM}-based controller similar to \citet{Zoph_2017_Neural}.
		To collapse the \ac{MORL} problem to a single-objective reinforcement learning problem, \citeauthor{Tan_2019_MnasNet} opted for a non-linear scalarization approach.
		They achieved this through the use of product scalarization, where the accuracy is multiplied with a value computed from the inference latency, adjusted so a specific latency threshold can be met.

		In \citeyear{Li_2023_GraphPNAS}, \citet{Li_2023_GraphPNAS} introduced GraphPNAS, a probabilistic graph generator.
		The generator is trained using the REINFORCE algorithm~\cite{Williams_1992_Simple} and evaluated on a variety of search spaces, including the RandWire search space, the \ac{ENAS} search space~\cite{Pham_2018_Efficient} and the NAS-Bench-101~\cite{Ying_2019_NASBench101} and -201 search spaces~\cite{Dong_2020_NAS}.
		GraphPNAS is shown to exhibit strong performance across all search spaces, outperforming the \ac{RNN}-based controller introduced by \citet{Zoph_2017_Neural} and commonly used up to that point.
		On the RandWire search space GraphPNAS requires 16 \acs{GPU}-days to complete the search with an oracle evaluator, while on the \ac{ENAS} macro search space, 12 \acs{GPU}-hours are required.

		\citet{Cassimon_2024_Scalable} recently introduced a novel reinforcement learning-based \ac{NAS} agent with the aim of building a reusable search behaviour.
		Their agent iteratively makes changes to a given architecture until the episode is terminated, either because the agent terminated it, or because a maximum episode length was reached.
		They evaluate their agent on the NAS-Bench-101 and NAS-Bench-301 benchmarks and demonstrate that their agent is capable of designing architectures competitive with those designed by other state-of-the-art algorithms.
		While the agent presented by \citet{Cassimon_2024_Scalable} offers strong performance, it still suffers from having to retrain the agent from scratch for every new task.
		This is particularly troublesome when training the agent takes dozens (NAS-Bench-101) or even hundreds (NAS-Bench-301) of GPU-hours.

	\subsection{Transfer Learning in Reinforcement Learning}
		In their survey paper, \citet{Zhu_2023_Transfer} cover many recent transfer learning techniques used in the field of reinforcement learning.
		They categorize transfer learning approaches primarily based on what knowledge is being transferred, arriving at five distinct categories: Reward Shaping, Learning from Demonstrations, Policy Reuse, Inter-Task Mapping and, Representation Transfer.
		Each of these five categories exhibits a number of properties particular to that category of transfer learning algorithm, and is usually evaluated on a specific set of metrics.

		In the context of this paper we will focus particular on fine-tuning, since that is the technique applied here.

		Fine-tuning is a transfer learning technique that involves adapting a policy that was pretrained on one task to another task.
		\citet{Julian_2021_Never} successfully apply fine-tuning in a reinforcement learning system for the task of grasping objects using a robot arm.
		They show that pretraining using reinforcement learning is essential, and allows their reinforcement learning policy to adjust to new tasks with relatively little data.
		\citet{Julian_2021_Never} consider several variations on the original grasping task, such as changing the geometry of the robot arm, or the lighting conditions observed by the camera.
		They also evaluate their fine-tuning methodology in a continual learning setting, and demonstrate strong results.
		% \todo{Kevin: Only one relevant reference}

	\subsection{Multi-Task \ac{NAS}}
		%As \ac{NAS} systems are becoming more and more advanced, some publications have started to consider a multi-task \ac{NAS} setting.
		%In this setting, the flexibility of \ac{NAS} systems is tested and improved, leading to \ac{NAS} systems that can design neural networks for multiple problem domains.
		%While many methodologies allow for easy adaptation to new, unseen tasks, the search space considered is often equally important in determining which tasks can be transferred to.

		%\todo{Kevin: In the multi-task NAS setting, do all tasks need to be known beforehand? If so, this is an important limitation that is not mentioned.}
		In their work, \citet{Ding_2022_Learning} use \ac{NCP} to find neural network architectures that are suitable for multiple tasks.
		They tackle this by specifically designing their network from the outset to be able to perform various tasks, rather than trying to transfer from one task to another post-hoc, as is often the case~\cite{Huh_2016_What,Qian_2023_Deep}.
		Their \ac{NCP} algorithm is able to traverse the search space they operate in by following the direction of the gradient of their predictor with respect to the network code given as the input to the network.
		The system naturally extends to the multitask setting where the set of tasks is known by accumulating gradients from multiple predictor heads.
		They consider two strategies for search space traversal: Continuous Propagation and Winner-Takes-All Propagation.
		Continuous Propagation updates all dimensions of the coding at the same time.
		Winner-Takes-All Propagation involves updating only the coding dimension with the highest gradient at each time step (thus possibly updating different dimensions in an alternating fashion).
		Besides their introduction of \ac{NCP}, \citeauthor{Ding_2022_Learning} also introduce NAS-Bench-MR, \iac{NAS} benchmark designed around a multi-branch, multi-resolution search space.
		They also evaluate their \ac{NCP} approach on NAS-Bench-MR.

		In their work, \citeauthor{Huang_2022_Arch} propose the use of pairwise rank prediction to find a partial ordering over architectures~\cite{Huang_2022_Arch}.
		Through the use of task embeddings and transfer learning by way of fine-tuning, their system is capable of easily adapting to unseen computer vision problems with limited amounts of additional training.
		They demonstrate their method using the Trans-NASBench-101 and NAS-Bench-201 benchmarks, and achieve results competitive with the state of the art on both benchmarks.
		One major shortcoming of their method is the fact that it relies on the use of ResNet50 to generate task embeddings, which inherently limits the set of tasks that ArchGraph can be applied to, to tasks that can be approached with ResNet50, excluding tasks such as language generation, which are often approached using transformer or \ac{LSTM}-based architectures.

		\Ac{AIOP} is a performance prediction algorithm proposed by \citet{Mills_2023_AIOP}.
		It is built on the use of \acp{GNN} and K-Adapters to easily transfer a pretrained performance predictor to a new task.
		They use a low-level computational graph which includes features like kernel sizes, strides and channel counts of a given architecture as input for their predictor.
		Based on this, they attempt to predict a generic performance score on a specific task, which is then scaled using a label scaling method to the appropriate performance value.
		In order to train their K-Adapters, \citet{Mills_2023_AIOP} make use of a shared head approach with latent sampling.
		This reduces the computational resources required to transfer their performance predictor to unseen tasks.
		While their method is very flexible, a new K-Adapter must be trained for every new task considered.

		\citet{He_2024_Robustifying} propose \iac{NAS} algorithm based on a combination of \ac{BO} and greedy search using a linear weighted combination of training-free metrics.
		They initially use \ac{BO} to gather a set of observations, and then use greedy search within this set of observations to find the best performing architecture.
		Additionally, \citet{He_2024_Robustifying} show that the performance of their algorithm can be theoretically guaranteed.
		The algorithm is evaluated on NAS-Bench-201, Trans-NASBench-101 and \ac{DARTS}, showing strong performance on all three benchmarks.
		Given that their algorithm relies exclusively on training-free metrics, it can very easily be adapted to novel tasks, assuming the underlying training-free metrics do not deteriorate significantly on these novel tasks.

		\citet{Zhou_2024_Toward} introduce \ac{MTNAS}, \iac{NAS} method that is designed to design neural networks for several different tasks concurrently.
		They achieve this by concurrently executing multiple evolutionary algorithms, one for each task, and exchanging knowledge between the different processes to speed up the search process.
		While the method does achieve efficiency gains through the concurrent evolution of architectures for multiple tasks, the set of tasks that are considered is fixed when the evolutionary algorithm is run.
		Their method is evaluated on the NAS-Bench-201 and Trans-NASBench-101 benchmarks, as well as on the \ac{DARTS} search space.
		Their findings indicate that their method is capable of finding architectures competitive with those found by state-of-the-art methods with a limited computational cost.
		\citet{Zhou_2024_Toward} also conduct ablation studies to show the added values of the various components of their methodology.

\section{Methods}
\label{sec:methods}
	In this publication, we aim to evaluate the flexibility of the reinforcement learning agent first published by \citet{Cassimon_2024_Scalable} in terms of task adaptation through transfer learning.
	We adopt one of the simplest possible forms of transfer learning.
	First, an agent is trained on one task after which the trained parameters are used as an initialization for training an agent on another task.
	This is repeated for every pair of tasks from the selected set of tasks.
	In the following sections we elaborate on some of the aspects of our methodology.
	Section~\ref{subsec:methods:task-selection} elaborates on the selection procedure used to determine a subset of tasks from the Trans-NASBench-101 benchmark that is used in this publication.
	Section~\ref{subsec:methods:reward-shaping} details the reward shaping mechanism used in this publication, since it differs from the one used by \citet{Cassimon_2024_Scalable}.

	\subsection{Task Selection}
	\label{subsec:methods:task-selection}
		\begin{figure}
			\centering
			\includegraphics[width=\columnwidth,clip]{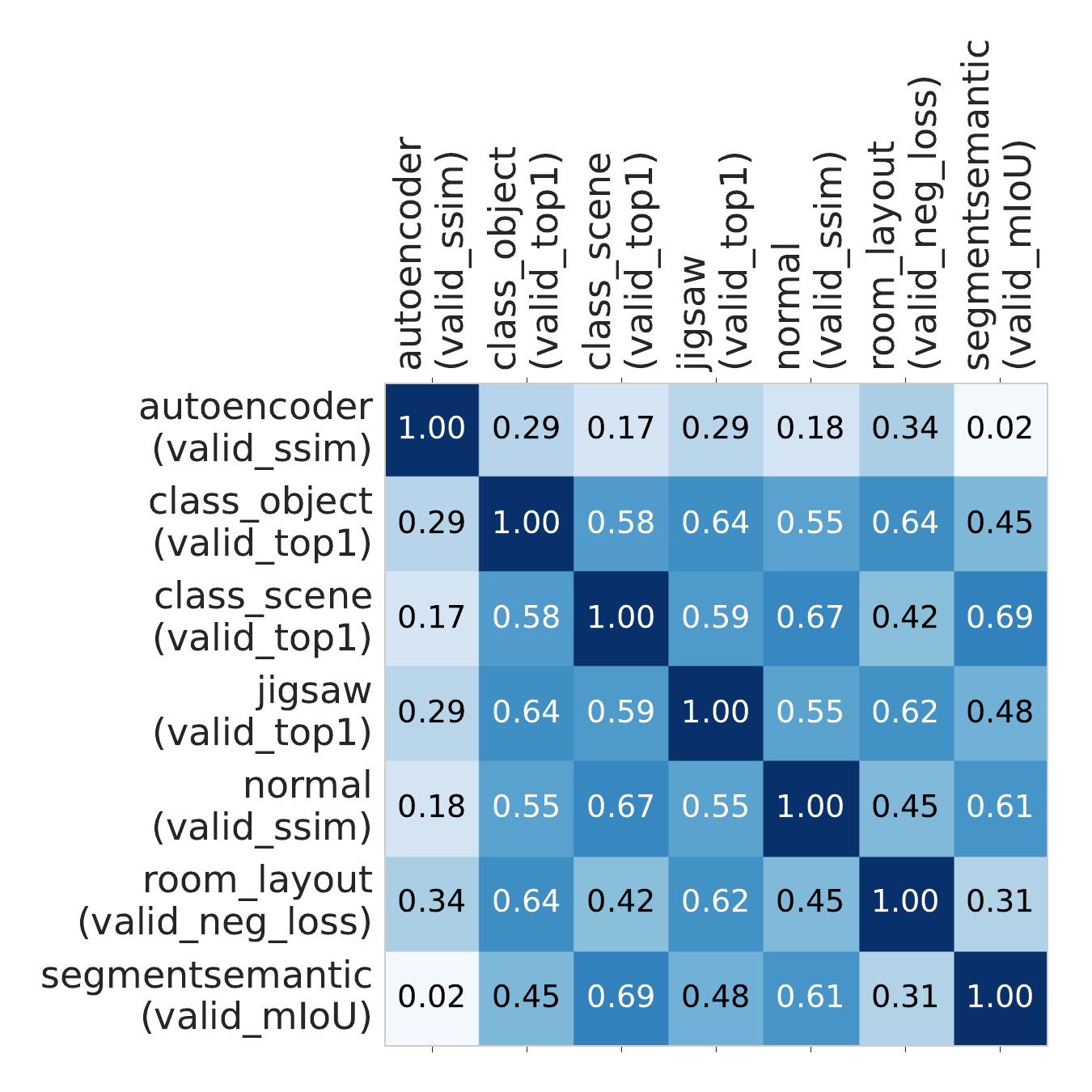}
			\caption{The Kendall's tau ranking correlation between the different tasks in the Trans-NASBench-101 benchmark.}
			\label{fig:tnb101-validation-correlation}
		\end{figure}

		% As mentioned before, we use the reinforcement learning agents introduced by \citet{Cassimon_2024_Scalable}, and place them in a transfer learning setting to assess the advantages and disadvantages of transfer learning to quickly adapt reinforcement learning agents to new tasks.
		To perform our assessment, we make use of the Trans-NASBench-101 benchmark~\cite{Duan_2021_TransNAS}.
		Since training an agent on every task in the Trans-NASBench-101 setting and transferring to every other task would be prohibitively expensive in terms of computational resources, we select a limited number of tasks to evaluate our agents on: object classification, room layout, autoencoding and semantic segmentation.
		We selected these tasks due to the variety between them, excluding scene classification and jigsaw (both classification problems) and surface normal (similar to autoencoding).
		As displayed in Figure~\ref{fig:tnb101-validation-correlation}, the correlation between the selected tasks is low, which should prove a challenge for our reinforcement learning agent.
		We used the same metrics as the original Trans-NASBench-101 paper: \ac{SSIM} for autoencoding tasks, top-1 accuracy for classification tasks, negative loss for room layout and mean \ac{IoU} for semantic segmentation.

	\subsection{Reward Shaping}
	\label{subsec:methods:reward-shaping}
		\begin{figure}
			\centering
			\includegraphics[width=\columnwidth,trim=0.5cm 0.8cm 0.8cm 0.8cm,clip]{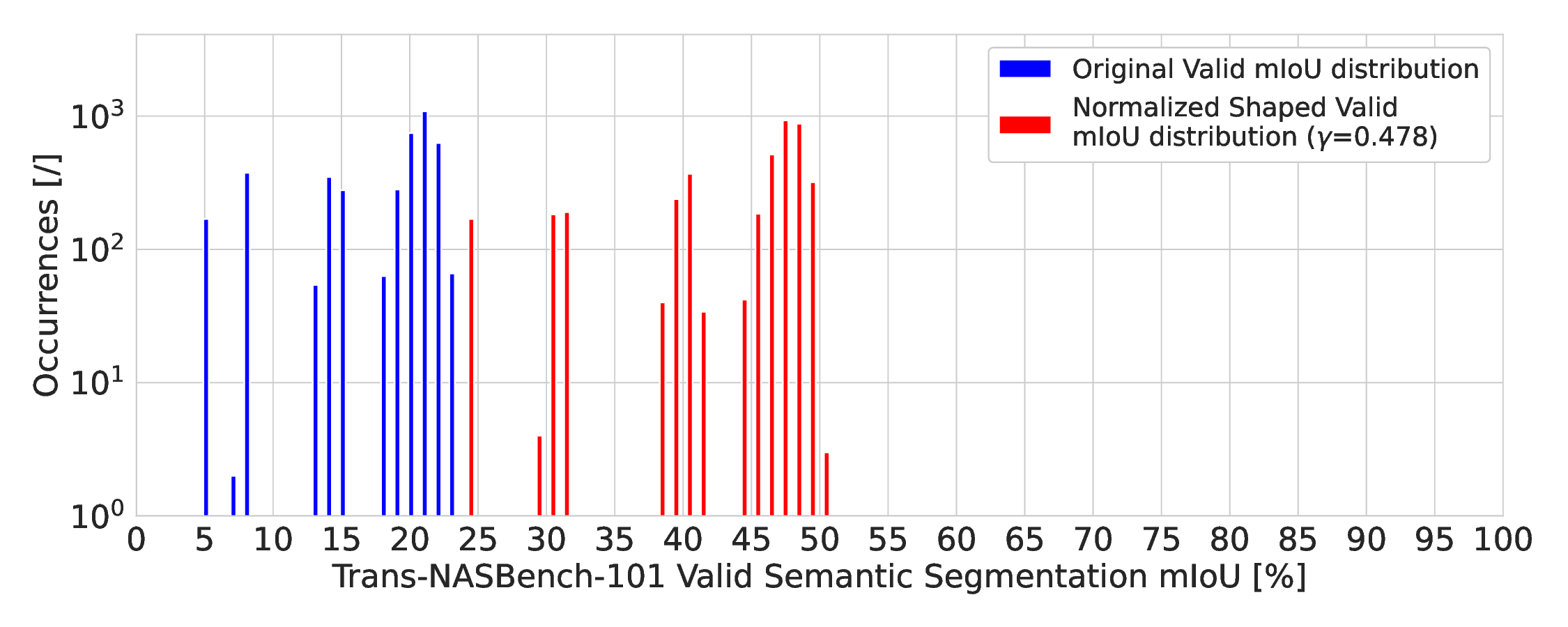}
			\caption{A comparison of the original and gamma-transformed reward distribution for the validation set for the segmentsemantic task.}
			\label{fig:segsem_histogram_comp}
		\end{figure}

\begin{figure}
			\centering
			\includegraphics[width=\columnwidth, clip]{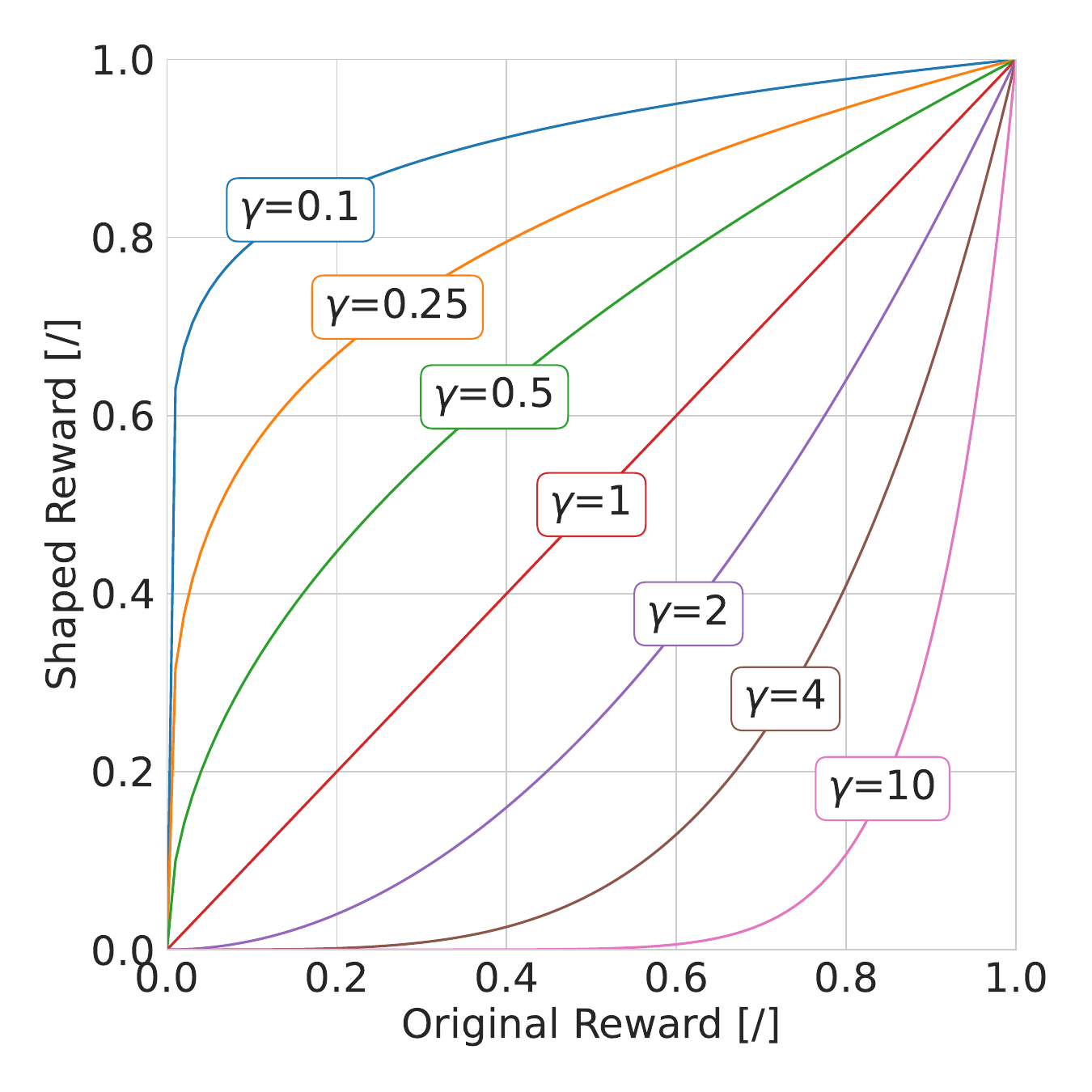}
			\caption{A comparison of our reward shaping function for different values of the $\gamma$ parameter.}
			\label{fig:segsem_reward_shaping:gamma_comparison}
		\end{figure}

\begin{figure*}
			\centering
			\includegraphics[width=\textwidth, trim=0.5cm 0.8cm 0.8cm 0.8cm, clip]{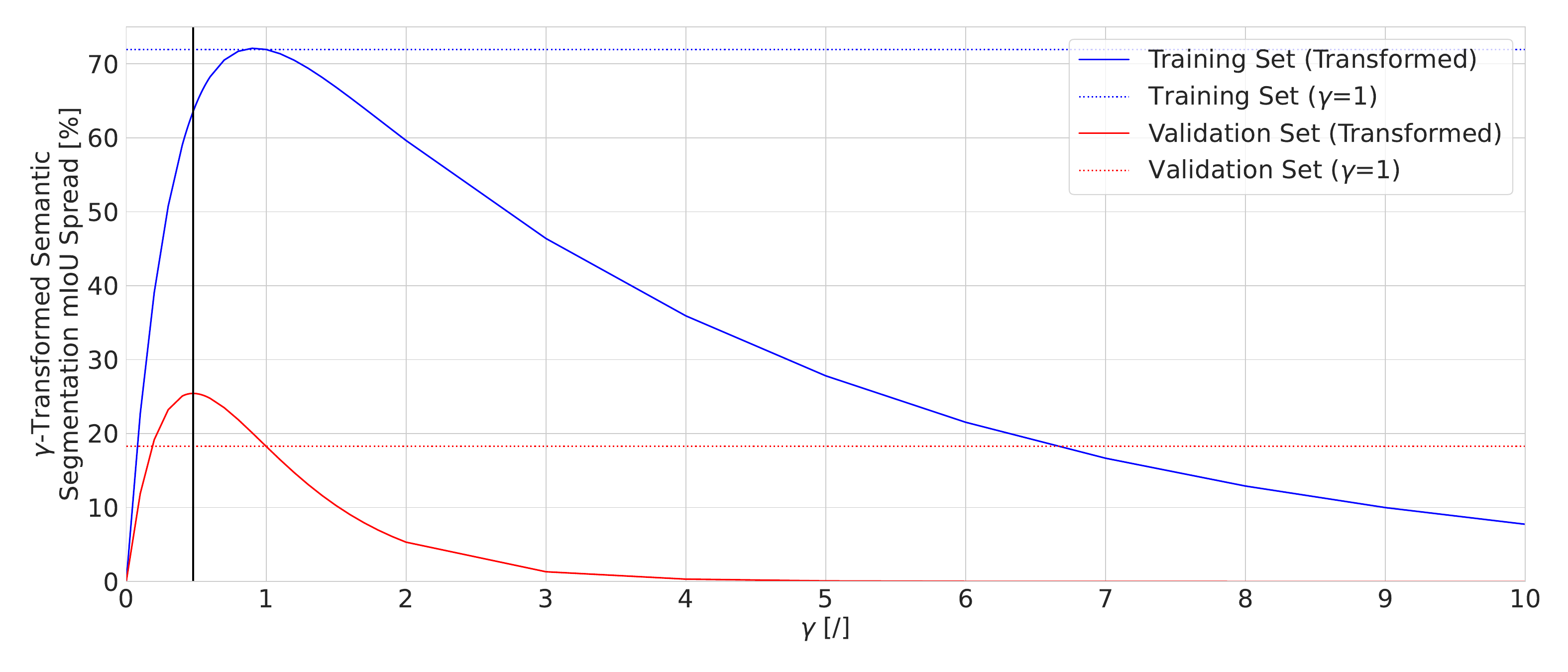}
			\caption{The spread in reward values for a given value of $\gamma$ in the gamma transform. The maximum validation spread is marked with a vertical solid black line.}
			\label{fig:segsem_reward_shaping:gamma_sweep}
		\end{figure*}

		During exploratory experiments, we noticed that, similar to \citet{Cassimon_2024_Scalable}, some of our reward functions have a sub-optimal distribution.
		In particular, the semantic segmentation mIoU has a very low spread and is concentrated at the lower end of its possible range (0-100\%), as demonstrated in Figure~\ref{fig:segsem_histogram_comp}.
		This results in very small reward values with a relatively small spread, and makes training a good policy more difficult.
		Unfortunately, the reward shaping employed in \citet{Cassimon_2024_Scalable} is not usable here, since it is designed to spread out high reward values across the whole range, whereas we need to spread out low reward values across the whole range.
		With this in mind, we introduce a new reward shaping mechanism based on the gamma transform: $R'\left(s, a\right) = R\left(s, a\right)^{\gamma}$.
		This mechanism is designed to spread out reward values on either end of the reward spectrum, depending on the choice of the parameter $\gamma$.
		Figure~\ref{fig:segsem_reward_shaping:gamma_comparison} shows a comparison of the $\gamma$-transform for different values of $\gamma$.
		Empirically, we find that for the semantic segmentation task, $\gamma=0.478$ is the optimal value.
		We showcase the spread of the rewards as a function of a sweep of $\gamma$ values in Figure~\ref{fig:segsem_reward_shaping:gamma_sweep}.

		Using the transform showcased in Figure~\ref{fig:segsem_reward_shaping:gamma_comparison}, we can improve the spread of the reward values and shift the reward values up across the entire range.
		% We also note that, in contrast to the original distribution, no single bin has more than $10^{3}$ samples out of a total of $4.096 \times 10^{3}$, showing that our reward shaping mechanism was effective at spre new reward distribution is more diffuse.\todo{Is diffuse the correct word?}

\section{Experiments}
\label{sec:experiments}
	The next section will detail the experiments we conducted to investigate the research questions posited earlier.
	In Section~\ref{subsec:experiments:pretraining-benefits}, we start by assessing if there is any benefit to pretraining on one task and transferring to another task.
	Next, in Section~\ref{subsec:experiments:transfer-regime-comparison}, we compare zero-shot transfer to a transfer regime with limited and full retraining.
	Finally, Section~\ref{subsec:experiments:required-additional-training} considers a regime of limited fine-tuning, and how much additional training is required to match or exceed the performance of an agent trained from scratch on a task.

	In the following sections, we will analyze the same set of experiments from different angles.
	To collect the data used in our analysis, first, one agent per random initialization was trained on each of the four tasks considered in this work.
	Next, the parameters of each of the four pretrained agents were used to initialize an agent trained on each of the three other tasks.
	In total, this yields 16 agents ($4 + \left(4 \times 3\right)$) per random initialization.

	Throughout our experiments, we considered three transfer learning regimes.
	The first regime is direct or zero-shot transfer.
	Under this regime, agents are trained for $1 \times 10^{7}$ time steps on the source task, and evaluated on the target task without any additional training.
	The second regime is the fine-tuning regime, where agents are trained for $1 \times 10^{7}$ time steps on the source task, followed by an additional $1 \times 10^{6}$ time steps on the target task.
	Following the training on the target task, the agents are evaluated on the target task.
	Finally, the third regime we consider is the re-training regime, where agents are first trained for $1 \times 10^{7}$ time steps on the source task, followed by another $1 \times 10^{7}$ time steps of training on the target task.
	After training on the target task, agents are evaluated on the target task.
	In cases where the source and target task are identical (Such as the elements on the main diagonal in Figure~\ref{fig:matrices:zero-shot}), agents are only trained on the task at hand for $1 \times 10^{7}$ time steps (i.e., the training procedure isn't repeated twice).
	The data used for the fine-tuning regime is the same data that was used for the retraining regime, but with the training procedure truncated after $1 \times 10^{6}$ time steps of training.

	The hyperparameters we use are mostly the same as those used by \citet{Cassimon_2024_Scalable}.
	Both when training on the source task, and when transferring to the target task, agents were trained for $10 \times 10^{6}$ time steps using the Ape-X algorithm in batches of $256$.
	Our agents use double Q-learning and dueling heads, along with 3-step bootstrapping enhanced using \ac{PEB}~\cite{Pardo_2018_Time}.
	We use a target network to stabilize training, updated every $8192$ time steps of training.
	Parameters were optimized using the Adam optimizer~\cite{Kingma_2014_Adam} with a learning rate of $5 \times 10^{-5}$.
	Gradients are clipped to an L2 norm of 40 during the optimization procedure.
	No reward shaping was used, except when the task being trained for was ``segmentsemantic'', in this case, the gamma transform with $\gamma=0.478$ was used, as detailed in Section~\ref{subsec:methods:reward-shaping}.
	Performance metrics on the validation set are used for training, while test set metrics are used for evaluation, unless otherwise noted.
	Similar to \citeauthor{Cassimon_2024_Scalable}, we used 32-fold environment vectorization, with 1 GPU and 2 CPU cores allocated to the driver.
	We used 8 workers with 1 CPU core each and no GPUs to gather experience, along with 4 replay buffer shards that were allocated 1 CPU core each.
	Our replay buffer is a prioritized replay buffer~\cite{Schaul_2016_prioritized} with a capacity of $25 \times 10^{3}$ entries, and $\alpha=0.6$, $\beta=0.4$, with initial sample priorities being computed on the experience collecting workers.
	The environment presents the agent with up to 50 neighbours at a time (plus the current architecture).
	Observation processing is the same as \citet{Cassimon_2024_Scalable}, adjacency matrices are padded randomly along the main diagonal before being prepared and given to the agent as input, with operations being one-hot encoded.
	Before being presented to the transformer encoder, architectures are embedded into a 256-dimensional latent space using three linear layers with ReLU activations.
	Our transformer consists of 2 layers with 4 heads each, and is followed by an advantage and value head of 3 linear layers with ReLU activations, each.

	\subsection{Benefit of Pretraining}
	\label{subsec:experiments:pretraining-benefits}
		\begin{figure}
			\centering
			\includegraphics[width=\columnwidth,trim=0.5cm 1.0cm 1.2cm 1.0cm,clip]{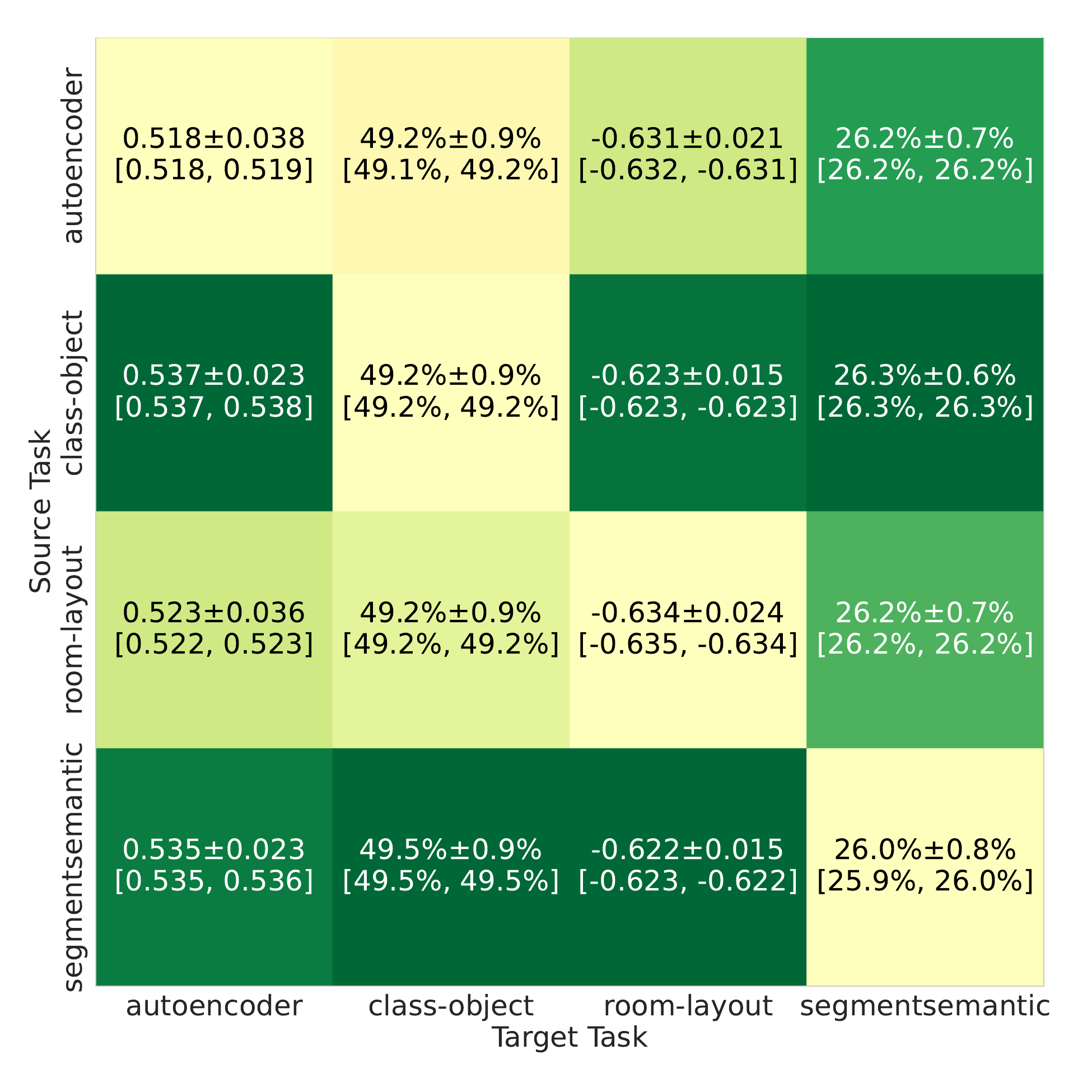}
			\caption{Performance of agents trained on one task and transferred to another under the retraining regime.}
			\label{fig:matrices:retrain}
		\end{figure}

\begin{figure}
			\centering
			\includegraphics[width=\columnwidth,trim=0.5cm 1.0cm 1.2cm 1.0cm,clip]{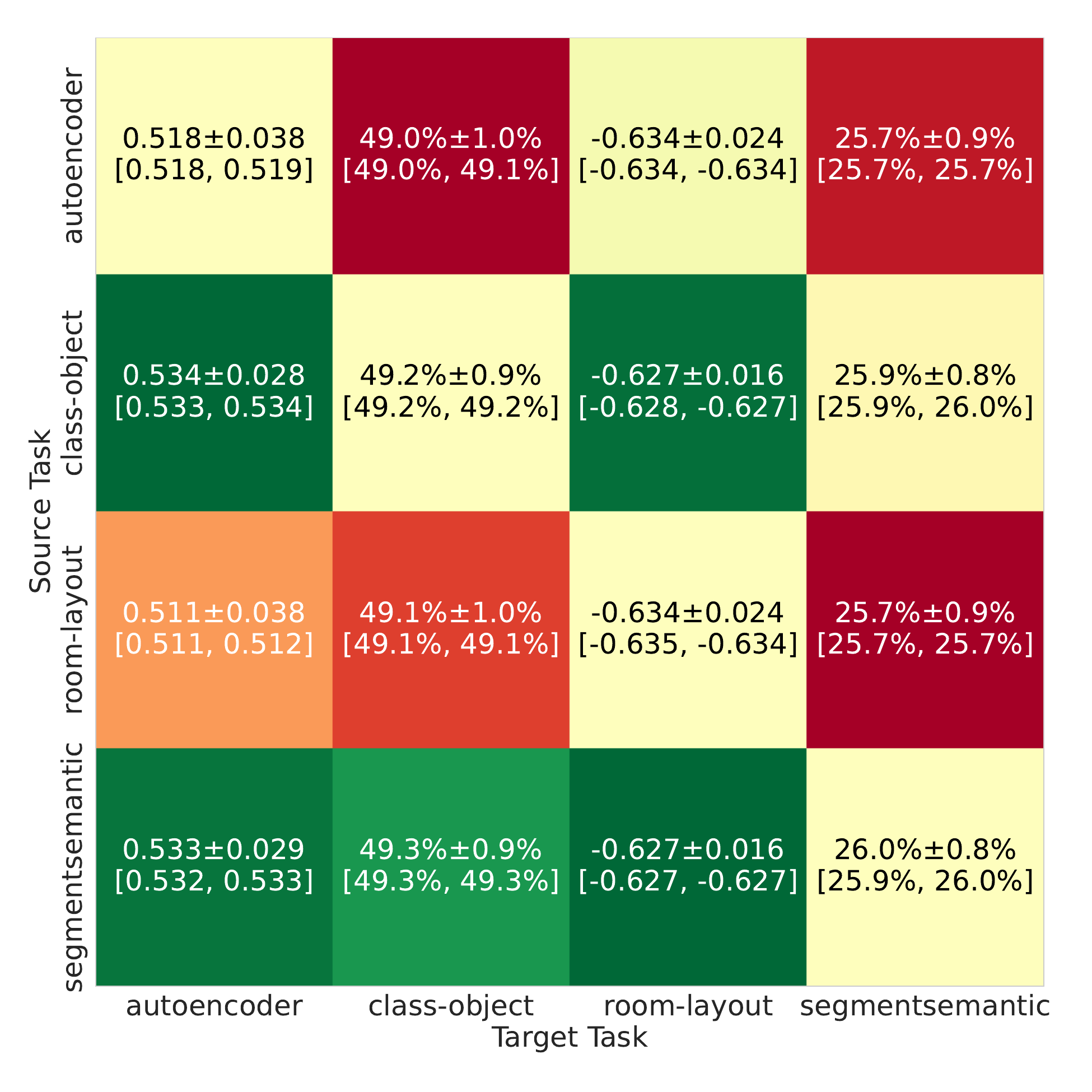}
			\caption{Performance of agents trained on one task and transferred to another under the fine-tuning transfer learning regime.}
			\label{fig:matrices:fine-tune}
		\end{figure}

		We first assess the benefits of pretraining by comparing the performance of agents trained from scratch on a task, to that of agents that were first trained on a different task and then transferred to that task.
		For this assessment, we will use the data from the retraining regime, visualized in Figure~\ref{fig:matrices:retrain}.
		Every element of the matrix shows the performance of one set of agents that was either trained from scratch for a task (main diagonal), or transferred from one task to another (off-diagonal elements).
		We record the performance on a specific task as the mean, the standard deviation and a 95\% confidence interval around the mean.
		Figure~\ref{fig:matrices:retrain} shows that in almost all cases, agents pretrained on a task and transferred to another task actually outperform agents trained from scratch on that task.
		In fact, the confidence interval of the agent trained from scratch only overlaps with the transferred agents when the target task is ``class-object'', in all other cases, there is no overlap in the confidence intervals of the agent trained from scratch for a task when compared to agents that were transferred to the task.
		A lack of overlap indicates that there is a statistically significant difference in performance between transferred agents and agents trained from scratch, with transferred agents outperforming agents trained from scratch.
		The same is also present under the fine-tuning regime showcased in Figure~\ref{fig:matrices:fine-tune}, where agents have only been trained on the target task for $1 \times 10^{6}$ time steps.

		We provide two possible hypotheses to explain this phenomenon.
		The first is that the process of finding optimal agent parameters is prone to getting stuck in local optima.
		This is what we observe when we train an agent from scratch, with the optimization process rediscovering the same (or similar) local optima between different random initializations.
		When initial conditions are changed by initializing the agent with a different set of pretrained weights, the distribution of experiences the agent collects also shifts, thus severely impacting the optimization procedure.
		In this case, the use of more advanced exploration strategies could be a viable path to improving the performance of agents trained from scratch.
		A second hypothesis is that our agents display a form of grokking~\cite{Power_2021_Grokking}.
		This is especially apparent when considering the results for room layout and semantic segmentation in Figure~\ref{fig:training-curves-comparison:10M}.
		The graphs show that long after the obtained performance of the agent has stabilized (2M time steps trained for both room layout and semantic segmentation), the agent can still make significant improvements.
		Under this hypothesis, the training of agents initialized with the parameters of other agents can simply be seen as a longer training procedure of a non-stationary problem.
		% Since the training procedure is essentially twice as long, we are much more likely to have seen all grokking events, thus no longer yielding significant improvements.\todo{Rewrite this}
		While the phenomenon observed in this publication is similar to grokking, it isn't grokking.
		Grokking is described as a jump in accuracy on a validation set, following long after a neural network has started overfitting on its training set.
		While we observe a similar jump in performance, reinforcement learning doesn't really distinguish between training and validation data, thus making it hard to pinpoint a point at which our agent starts overfitting.
		Because the phenomenon we observe is slightly different, it is unclear if some of the mitigation measures touched on by \citet{Power_2021_Grokking} such as weight decay would be effective in our case.

	\subsection{Transfer Regime Comparison}
	\label{subsec:experiments:transfer-regime-comparison}
		\begin{figure}
			\centering
			\includegraphics[width=\columnwidth,trim=0.5cm 1.0cm 1.2cm 1.0cm,clip]{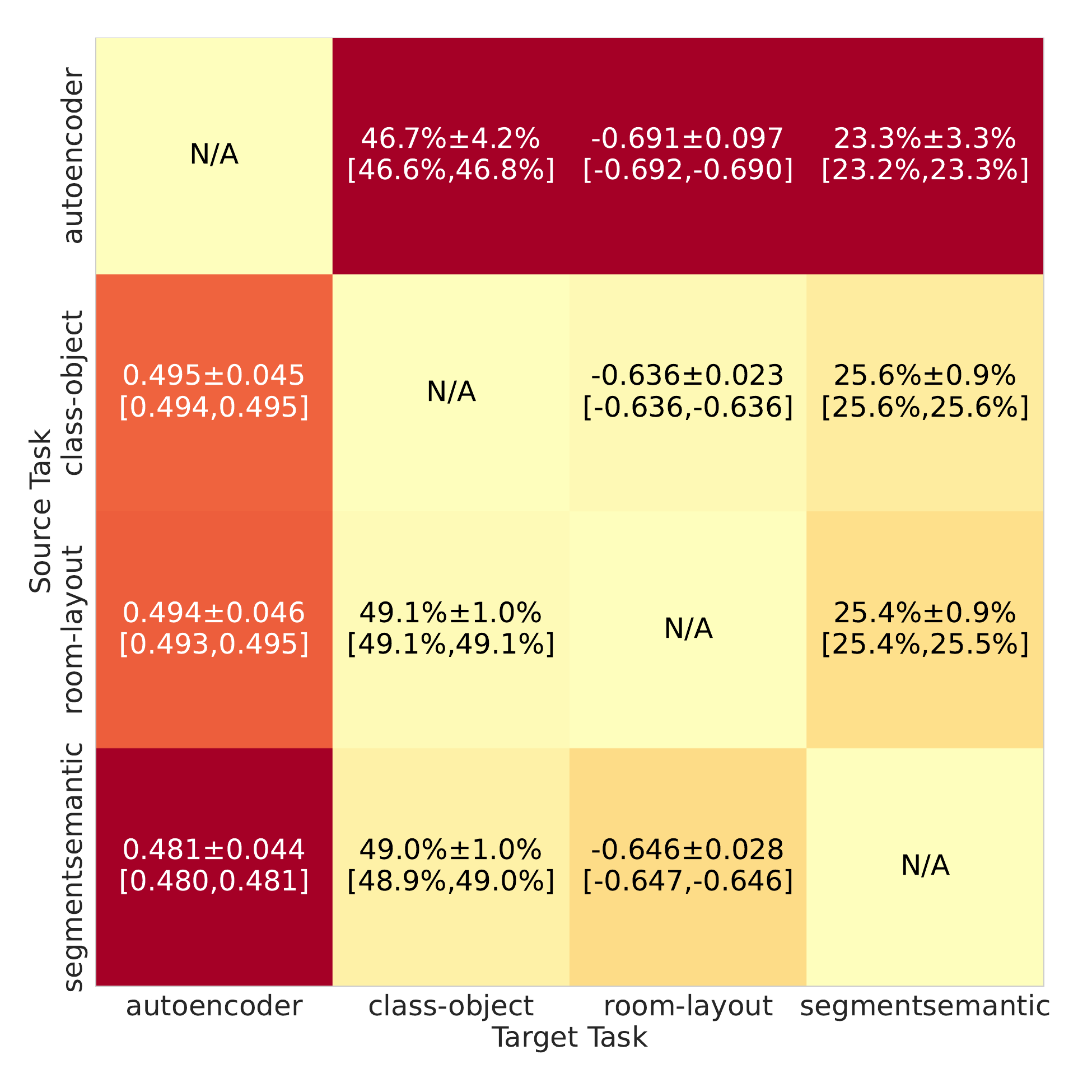}
			\caption{Performance of agents trained on one task and transferred to another under the zero-shot regime.}
			\label{fig:matrices:zero-shot}
		\end{figure}

		In this section, we consider the three different transfer learning regimes mentioned at the beginning of Section~\ref{sec:experiments}: zero-shot transfer, fine-tuning and retraining.

		Figures~\ref{fig:matrices:retrain},~\ref{fig:matrices:fine-tune} and~\ref{fig:matrices:zero-shot} show the performance of the reinforcement learning agent under different transfer learning regimes.
		Each cell in a matrix shows the mean, standard deviation and a 95\% confidence interval around the mean computed using bootstrapping.

		% \todo{Kevin: I am a bit undecided if this section (a) should remain as is, (b) be merged with the previous section, or (c) put before and limiting it to the zero-shot scenario (which is then used to motivate fine-tuning and lead to the next experiments).  The only new addition here are the zero shot results.}
		Comparing the matrices in Figure~\ref{fig:matrices:retrain} and Figure~\ref{fig:matrices:fine-tune} with the matrix in Figure~\ref{fig:matrices:zero-shot} shows the zero-shot transfer learning regime has the weakest performance.
		Figure~\ref{fig:matrices:zero-shot} also corroborates what we noted about Figure~\ref{fig:tnb101-validation-correlation}: The autoencoding task has a very low correlation with the other tasks, thus, an agent that was trained for the autoencoding task will perform poorly on other tasks under a zero-shot transfer learning regime.

		Comparing the zero-shot transfer results from Figure~\ref{fig:matrices:zero-shot} with the results from the agents that were fine-tuned in Figure~\ref{fig:matrices:fine-tune}, we can see that even a limited amount of fine-tuning can have a significant positive impact on the performance of each agent.
		In fact, Figure~\ref{fig:matrices:fine-tune} shows that even after fine-tuning for only $1 \times 10^{6}$ time steps, the agents trained from scratch are already outperformed by every transferred agent, regardless of the task it was originally trained for.
		Figure~\ref{fig:matrices:fine-tune} also shows that agents pretrained on the ``class-object'' and ``segmentsemantic'' tasks tend to outperform agents trained on the ``autoencoder'' and ``room-layout'' tasks.
		This too is corroborated by the training curves in Figure~\ref{fig:training-curves-comparison:10M} and~\ref{fig:training-curves-comparison:2M}, where agents pretrained on either of these tasks tend to be among the best performers.

	\subsection{Required amount of training}
	\label{subsec:experiments:required-additional-training}
		\begin{figure}
			\centering
			\includegraphics[width=\columnwidth, clip]{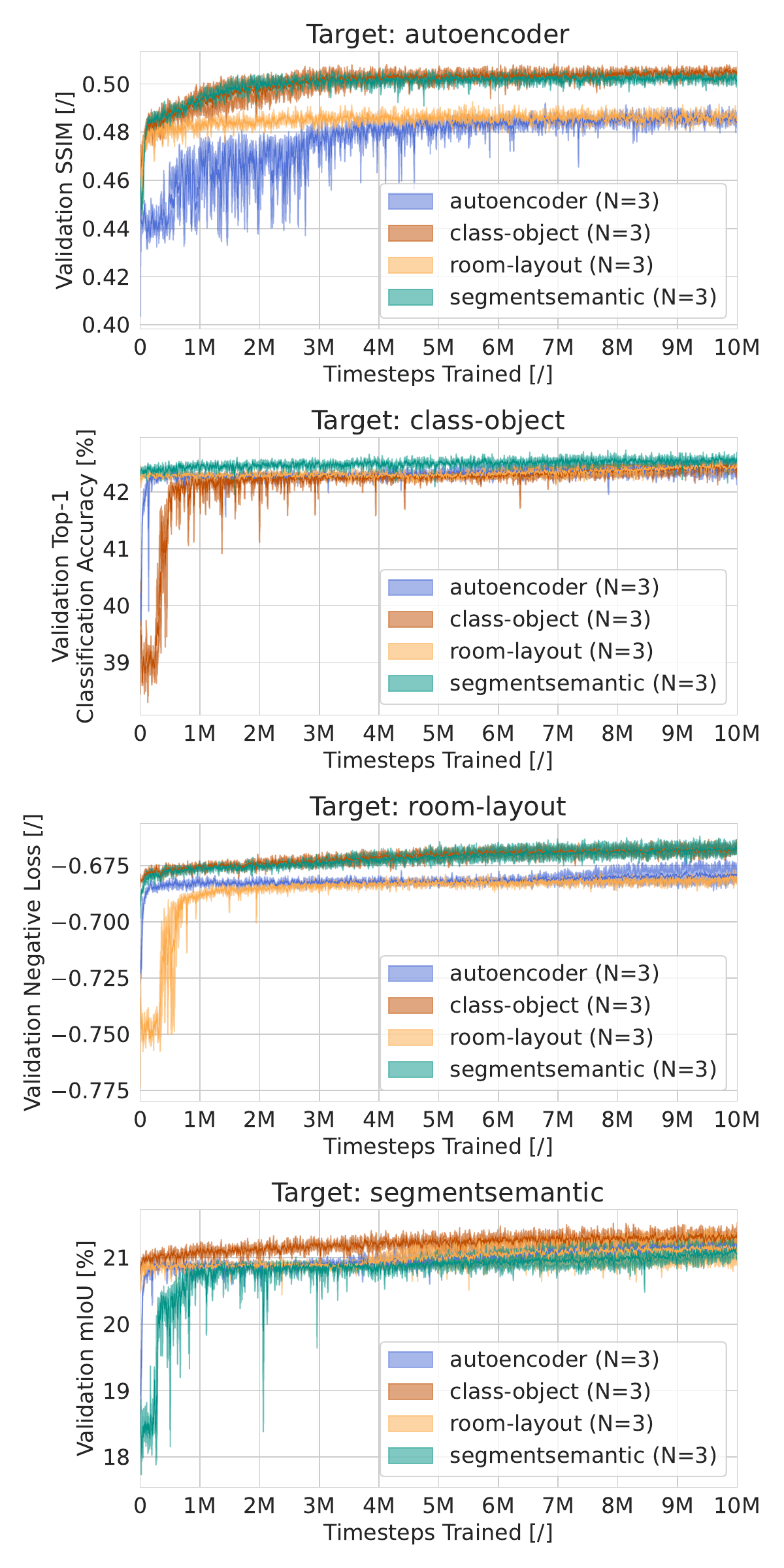}
			\caption{Training curves comparing train-from-scratch agents to agents being transferred from other problems.}
			\label{fig:training-curves-comparison:10M}
		\end{figure}

		\begin{figure}
			\centering
			\includegraphics[width=\columnwidth, clip]{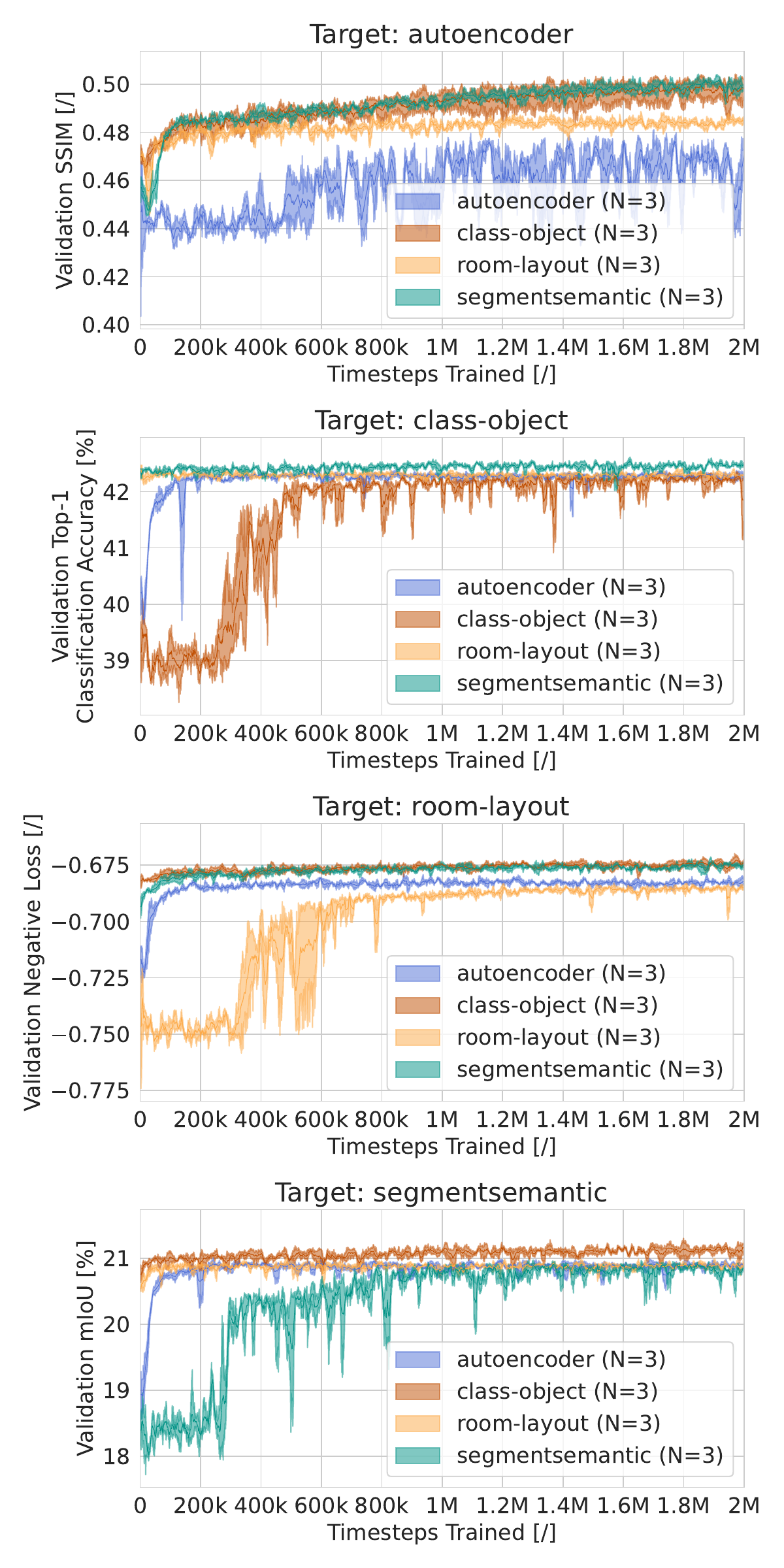}
			\caption{Training curves for the first 2M timesteps of training comparing train-from-scratch agents to agents being transferred from other problems.}
			\label{fig:training-curves-comparison:2M}
		\end{figure}

		In this section, we will attempt to quantify the amount of additional training required to transfer an agent from one domain to another, and achieve a performance similar to that of an agent trained from scratch.
		We can visualize this by comparing training curves, shown in Figure~\ref{fig:training-curves-comparison:10M} and~\ref{fig:training-curves-comparison:2M}.
		To compensate for the noisiness of the performance measurements, we filter each training curve with a simple moving average filter with a kernel size of 16 first.
		Each graph groups agents by target task following transfer from all source tasks.
		% \todo{Any idea on the performance if they would have been trained for 20M steps? Are they still outperformed by the transferred agents?} No.
		Lines indicate the sample mean, with the colored area representing a 95\% confidence interval around the mean.
		These training curves show that training an agent that has been pretrained on a different task is generally quicker than training an agent from scratch for almost any task.

		One of the main reasons for the quick training of the pretrained agents is likely that all agents operate in the same search space.
		The first layers of the neural network, which take an architecture and embed it into a latent space before feeding the latent space embedding to a transformer, could likely share a latent space across all problems without impacting task performance.
		Thus, while the transformer layers and all layers following them need to be retrained, it is likely that the first layers in the neural network remain relatively unchanged while transferring, thus aiding in speeding up convergence.

		We'd also like to note the length of the training procedure, in particular compared to the results obtained by \citet{Cassimon_2024_Scalable}.
		% \todo{Not sure if I follow here. Trans-NASBench, which is used in this study, has a smaller search space, but still requires the same amount of training as the larger search spaces in NAS-Bench-101 and NAS-Bench-301. Why is this a good thing? I would expect less training to be required? Looking at it in the opposite way, the larger search spaces don't need more training is indeed good. Is that what you are trying to say here?}
		Despite the fact that Trans-NASBench-101 has a relatively small search space ($4.096 \times 10^{3}$ compared to $4.23 \times 10^{5}$ for NAS-Bench-101 and $1 \times 10^{18}$ for NAS-Bench-301), it still requires roughly the same amount of training as the much larger problems studied by \citet{Cassimon_2024_Scalable}.
		This shows that the size of the search space has little effect on the convergence speed of the agent, underlining the scalability highlighted by \citet{Cassimon_2024_Scalable}.
		% This is particularly interesting in the light of the conclusion drawn by \citet{Cassimon_2024_Scalable} that their agent scales well with the search space, because it requires roughly the same amount of training for NAS-Bench-101 and NAS-Bench-301.
		While \citet{Cassimon_2024_Scalable} only provided two points of comparison, we add a third standardized \ac{NAS} benchmark to this, showing that the scalability demonstrated when going from NAS-Bench-101 to NAS-Bench-301 is also present when going from Trans-NASBench-101 to NAS-Bench-101.
		% Our results show that this relative inflexibility in the amount of training required also seems to extend to search spaces smaller than NAS-Bench-101, confirming the strong scaling properties of the \ac{NAS} agent with respect to the size of the search space.

		\begin{figure*}
			\centering
			\includegraphics[width=\textwidth, clip]{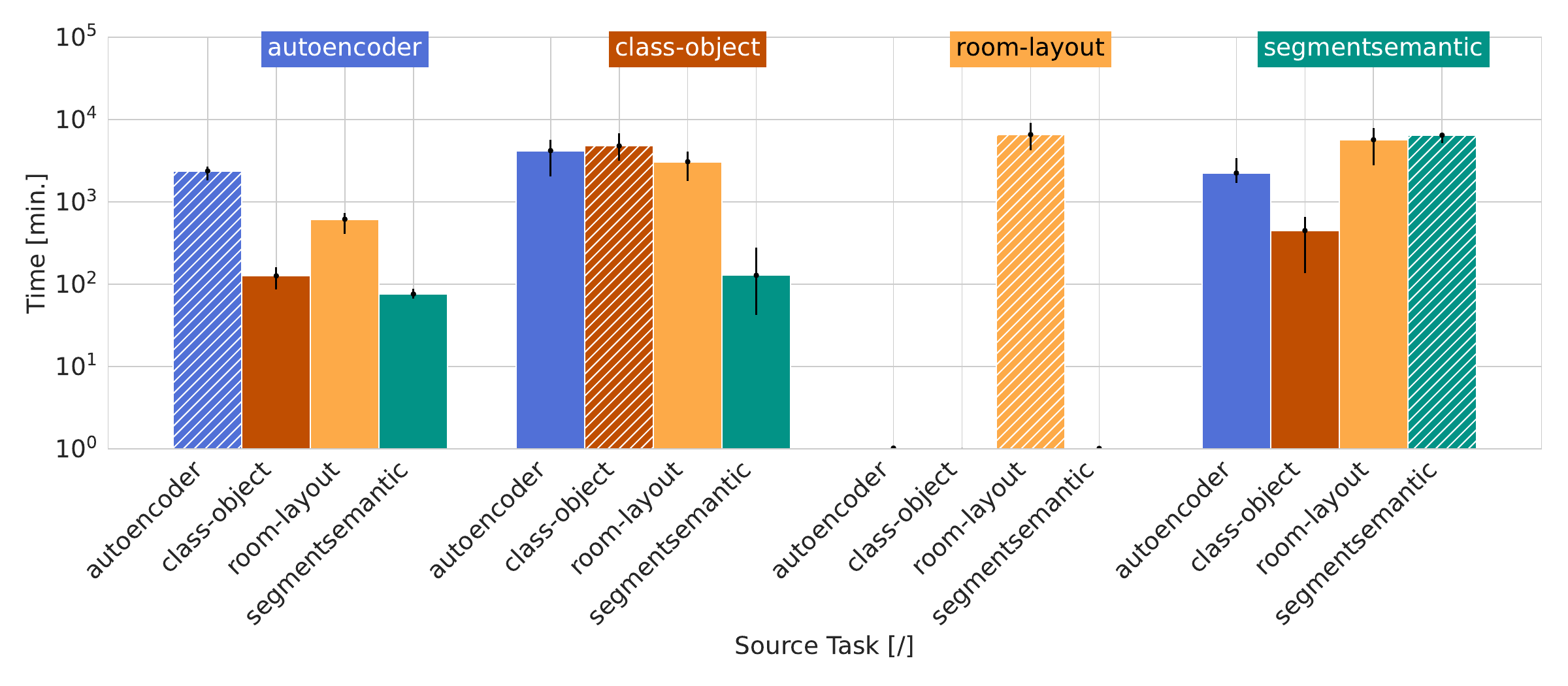}
			\caption{The time it takes agents to achieve performance equivalent to an agent trained froms scratch. Hatched bars indicate bars that show the time for agents trained from scratch. Error bars show 95\% confidence intervals. Bars are grouped by target task, with each bar representing a different source task. Time is given in wall-time.}
			\label{fig:time-to-equiv-perf:walltime}
		\end{figure*}

		\begin{figure*}
			\centering
			\includegraphics[width=\textwidth, clip]{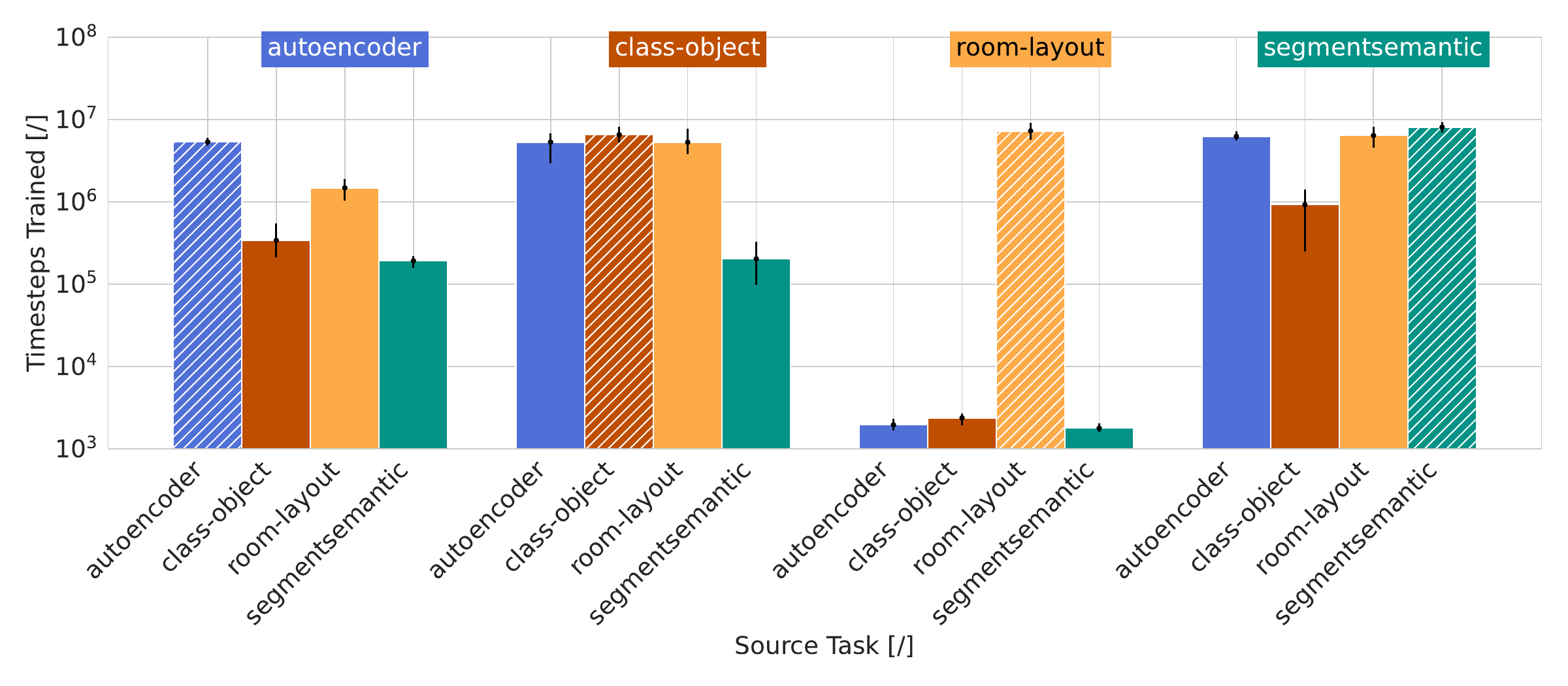}
			\caption{The time it takes agents to achieve performance equivalent to an agent trained from scratch. Hatched bars indicate bars that show the time for agents trained from scratch. Error bars show 95\% confidence intervals around the mean. Bars are grouped by target task, with each bar representing a different source task. Time is given in time steps.}
			\label{fig:time-to-equiv-perf:timesteps}
		\end{figure*}

		\begin{figure}
			\centering
			\includegraphics[width=\columnwidth, clip]{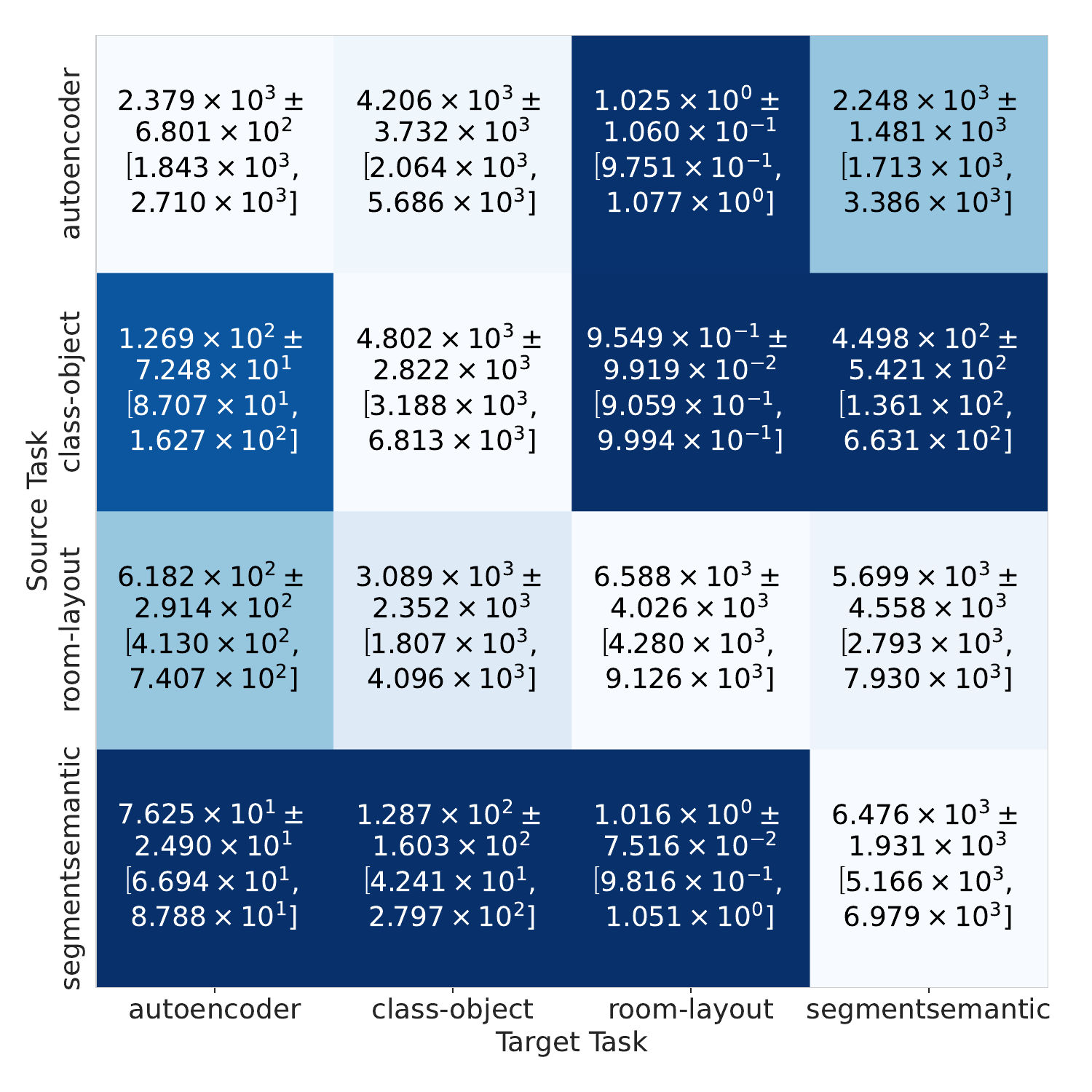}
			\caption{The time it takes agents to achieve performance equivalent to an agent trained froms scratch, expressed in minutes. Data is given as mean $\pm$ standard deviation along with a 95\%-CI around the mean.}
			\label{fig:time-to-equiv-perf:walltime:matrix}
		\end{figure}

		\begin{figure}
			\centering
			\includegraphics[width=\columnwidth, clip]{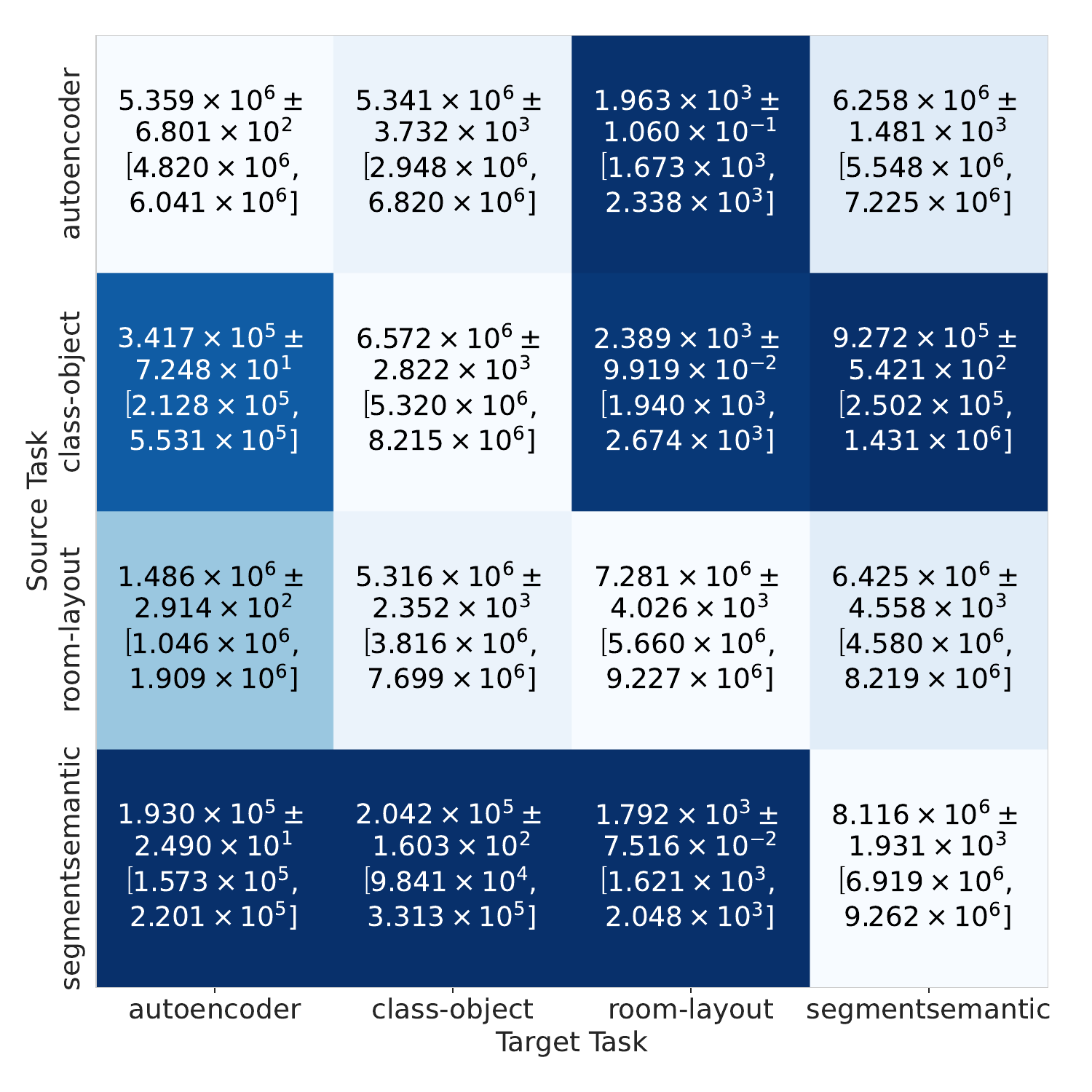}
			\caption{The time it takes agents to achieve performance equivalent to an agent trained froms scratch, expressed in timesteps. Data is given as mean $\pm$ standard deviation along with a 95\%-CI around the mean.}
			\label{fig:time-to-equiv-perf:timesteps:matrix}
		\end{figure}

		We can also consider the length of the training procedure from a different perspective.
		This is showcased in Figures~\ref{fig:time-to-equiv-perf:walltime} and~\ref{fig:time-to-equiv-perf:timesteps}, and numerically in Figure~\ref{fig:time-to-equiv-perf:walltime:matrix} and~\ref{fig:time-to-equiv-perf:timesteps:matrix}.
		We measure the amount of time needed, in terms of both wall-time and the number of time steps trained, to achieve the same performance as an agent trained from scratch for $10^{7}$ time steps.
		The bars are grouped based on their target task, and each bar represents a different source task.
		Hatched bars correspond to data for agents trained from scratch, while bars that aren't hatched correspond to agents that were transferred from one task to another.
		Metrics were computed using an all-to-all comparison scheme.
		Every training run that transferred from task A to task B was compared to every reference (where agents are trained from scratch for task B) run.
		The point at which the transferred runs start outperforming the runs trained from scratch was noted down for each pair of runs.
		Finally, a sample mean and a 95\% confidence interval were computed around the cross-over points, and shown on the graph.
		The height of each bar corresponds to the sample mean, while the error bar indicates the 95\% confidence interval.

		In a number of cases, there is no overlap in the 95\% confidence intervals between various task pairs, indicating a statistically significant difference in the amount of training necessary to match the reference performance.
		% \todo{Can you provide an explanation for these results.} I cannot
		When the target task is autoencoder or room layout, all agents that are transferred from other problems outperform the agents trained from scratch to a statistically significant degree.
		For the class object task, only the agents pretrained on the semantic segmentation task offer a statistically significant improvement in training time.
		In the case of the semantic segmentation task, agents pretrained on autoencoder and object classification provide a significant improvement in training time, while agents pretrained on room layout don't.

		For the room layout task, the contrast between pretraining and transferring is very large.
		This is also evident from the training curves showcased in Figure~\ref{fig:training-curves-comparison:10M} and~\ref{fig:training-curves-comparison:2M}.
		While the other three tasks still see a rise in performance relatively late in the training procedure, room layout performance flattens out rather quickly.

\section{Discussion and Conclusions}
\label{sec:discussion}
	In this publication, we set out to investigate the effectiveness of basic transfer learning routines to transfer a reinforcement learning-based \ac{NAS} agent from one task to another.
	We showed that even very basic transfer learning paradigms can allow agents to be effectively transferred between target tasks.
	While this varied between source and target tasks, in most cases, agents benefited from being pretrained on another task, with benefits being visible both in terms of a reduced training time on the target task, and better performance in the target task.
	Based on these observations, we conclude that reinforcement learning-based \ac{NAS} agents can be transferred between tasks, even with very basic transfer learning regimes.
	We hypothesize that more advanced transfer learning strategies, such as those detailed by \citet{Zhu_2023_Transfer} would likely result in even stronger performing agents.
	Another point of note is that this publication only considered a single reinforcement learning algorithm: Ape-X.
	The results we obtained are fairly convincing, but with the results presented here, it is impossible to ascertain if our conclusions also hold for other reinforcement learning algorithms.
	Despite these strong results we obtained, unlocking the full potential of such \ac{NAS} systems still requires further research.
	In the following sections, we set out some possible directions for future research.

	\subsection{Task Heterogeneity}
		In this paper, we selected four out of seven tasks from the Trans-NASBench-101~\cite{Duan_2021_TransNAS} paper to evaluate our methodology.
		Other \ac{NAS} benchmarks exist that contain performance data for different tasks, like NAS-Bench-201~\cite{Dong_2020_NAS}.
		Regardless, both Trans-NASBench-101 and NAS-Bench-201 are both limited to computer vision tasks.
		Other problem domains, like reinforcement learning, language modelling, etc.\ are completely absent from these benchmarks.

		In order to fully assess the ability of our methodology to transfer to a wide range of tasks, further experiments on other benchmarks such as NAS-Bench-360~\cite{Tu_2022_NASBench360} are necessary.

	\subsection{Search Space Generalization}
		This publication assumed that all tasks share the same search space.
		While this is viable when transferring between similar tasks, it may not be the case when a more diverse set of tasks is considered.

		One possible mitigation strategy for our agent to be able to deal with architecture being sampled from different search spaces is by pretraining the architecture encoding network in an autoencoder fashion.
		Training such an autoencoder only requires knowledge of the architectures being considered, without needing access to any performance data, sidestepping the costly process of gathering this performance data.
		The parameters of the architecture encoding network could then be frozen, and re-used when training agents for different tasks, similar to the bodyswapping methodology detailed by \citet{Mills_2023_AIOP}.

	\subsection{Multi-Task Reinforcement Learning}
		% \todo{Multi-task RL may also lead to better (more general) representations, which may further improve transfer learning.}
		In this publication, we considered only transfer learning methodologies that yield architectures well-suited to perform a single task.
		Given its simplicity, this approach allows us to effectively study the characteristics of our agent in detail.
		A more effective approach could be to make use of multi-task reinforcement learning techniques.
		In multi-task reinforcement learning, agents are designed to be capable of accomplishing multiple tasks in an optimal fashion.
		Using multi-task reinforcement learning, \ac{NAS} agents could be designed for several of the most common \ac{NAS} problems, requiring only a single training procedure to be executed.
		As an added benefit, multi-task reinforcement learning may also lead the agent to learn more general representations of the architectures it is given, leading to possible improvements in transfer learning performance.

	\subsection{Sample Efficiency}
		While transfer learning can help to speed up the process of training a reinforcement learning-based \ac{NAS} agent for a new task, this doesn't address a fundamental weakness of our agent.
		Ape-X is a reinforcement learning algorithm designed for improving the throughput of experience collection.
		Unfortunately, this comes at the cost of sample efficiency.
		While our agents are trained on $10 \times 10^{6}$ time steps of experience, it is not uncommon for $20 \times 10^{6} - 30 \times 10^{6}$ time steps of experience to be collected during the training procedure.
		This significantly raises the computational cost of training our reinforcement learning agent, and is a major flaw of the methodology.
		This could be mitigated by experimenting with other reinforcement learning algorithms with better sample efficiency.
		Model-based reinforcement learning is a particularly interesting avenue for future research, given the relative simplicity of the transition model in our \ac{MDP}.
		The transition model is trivial to model with perfect accuracy, thus allowing agents to learn much quicker and more effectively by utilizing this additional knowledge.

\backmatter

%\bmhead{Supplementary information}
%
%If your article has accompanying supplementary file/s please state so here.
%
%Authors reporting data from electrophoretic gels and blots should supply the full unprocessed scans for key as part of their Supplementary information. This may be requested by the editorial team/s if it is missing.
%
%Please refer to Journal-level guidance for any specific requirements.

\section*{Declarations}

% textidote: ignore begin
	\subsection{Data Availability}
% textidote: ignore end
		Datasets generated for this publication are available from the corresponding author upon reasonable request.
		Data for Trans-NASBench-101~\cite{Duan_2021_TransNAS} is available from the corresponding Github repository \url{https://github.com/yawen-d/TransNASBench}.

	\subsection{Funding}
		\acknowledgement

	\subsection{Conflicts of Interest}
		The authors have no relevant interests to disclose.

\bibliography{references}

\begin{thebibliography}{33}
\providecommand{\natexlab}[1]{#1}
\providecommand{\url}[1]{{#1}}
\providecommand{\urlprefix}{URL }
\providecommand{\doi}[1]{\url{https://doi.org/#1}}
\providecommand{\eprint}[2][]{\url{#2}}
 \bibcommenthead

\bibitem[{Baker et~al(2017)Baker, Gupta, Naik, and
  Raskar}]{Baker_2017_Designing}
Baker B, Gupta O, Naik N, et~al (2017) Designing neural network architectures
  using reinforcement learning. In: International Conference on Learning
  Representations, \urlprefix\url{https://openreview.net/forum?id=S1c2cvqee}

\bibitem[{Cassimon et~al(2024)Cassimon, Mercelis, and
  Mets}]{Cassimon_2024_Scalable}
Cassimon A, Mercelis S, Mets K (2024) Scalable reinforcement learning-based
  neural architecture search. Neural Computing and Applications
  \doi{10.1007/s00521-024-10445-2},
  \urlprefix\url{https://doi.org/10.1007/s00521-024-10445-2}

\bibitem[{Devlin et~al(2019)Devlin, Chang, Lee, and
  Toutanova}]{Devlin_2019_BERT}
Devlin J, Chang MW, Lee K, et~al (2019) {BERT}: Pre-training of deep
  bidirectional transformers for language understanding. In: Burstein J, Doran
  C, Solorio T (eds) Proceedings of the 2019 Conference of the North {A}merican
  Chapter of the Association for Computational Linguistics: Human Language
  Technologies, Volume 1 (Long and Short Papers). Association for Computational
  Linguistics, Minneapolis, Minnesota, pp 4171--4186,
  \doi{10.18653/v1/N19-1423}, \urlprefix\url{https://aclanthology.org/N19-1423}

\bibitem[{Ding et~al(2022)Ding, Huo, Lu, Yang, Wang, Lu, Wang, and
  Luo}]{Ding_2022_Learning}
Ding M, Huo Y, Lu H, et~al (2022) Learning versatile neural architectures by
  propagating network codes. In: International Conference on Learning
  Representations, \urlprefix\url{https://openreview.net/forum?id=KEQl-MZ5fg7}

\bibitem[{Dong and Yang(2020)}]{Dong_2020_NAS}
Dong X, Yang Y (2020) {NAS-Bench-201: Extending the Scope of Reproducible
  Neural Architecture Search} pp 1--16.
  \urlprefix\url{http://arxiv.org/abs/2001.00326}

\bibitem[{Duan et~al(2021)Duan, Chen, Xu, Chen, Liang, Zhang, and
  Li}]{Duan_2021_TransNAS}
Duan Y, Chen X, Xu H, et~al (2021) Transnas-bench-101: Improving
  transferability and generalizability of cross-task neural architecture
  search. In: Proceedings of the IEEE/CVF Conference on Computer Vision and
  Pattern Recognition (CVPR), pp 5251--5260

\bibitem[{Elsken et~al(2019)Elsken, Metzen, and Hutter}]{Elsken_2019_Efficient}
Elsken T, Metzen JH, Hutter F (2019) Efficient multi-objective neural
  architecture search via lamarckian evolution. In: International Conference on
  Learning Representations,
  \urlprefix\url{https://openreview.net/forum?id=ByME42AqK7}

\bibitem[{He et~al(2016)He, Zhang, Ren, and Sun}]{He_2016_Deep}
He K, Zhang X, Ren S, et~al (2016) Deep residual learning for image
  recognition. In: Proceedings of the IEEE Conference on Computer Vision and
  Pattern Recognition (CVPR)

\bibitem[{He et~al(2024)He, Shu, Dai, and Low}]{He_2024_Robustifying}
He Z, Shu Y, Dai Z, et~al (2024) Robustifying and boosting training-free neural
  architecture search. In: The Twelfth International Conference on Learning
  Representations, \urlprefix\url{https://openreview.net/forum?id=qPloNoDJZn}

\bibitem[{Huang et~al(2022)Huang, Huang, Li, Chen, Xu, Li, and
  Liang}]{Huang_2022_Arch}
Huang M, Huang Z, Li C, et~al (2022) Arch-graph: Acyclic architecture relation
  predictor for task-transferable neural architecture search. In: Proceedings
  of the IEEE/CVF Conference on Computer Vision and Pattern Recognition (CVPR),
  pp 11881--11891

\bibitem[{Huh et~al(2016)Huh, Agrawal, and Efros}]{Huh_2016_What}
Huh M, Agrawal P, Efros AA (2016) What makes imagenet good for transfer
  learning? CoRR abs/1608.08614.
  \urlprefix\url{http://arxiv.org/abs/1608.08614},
  {\href{https://arxiv.org/abs/1608.08614}{{1608.08614}}}

\bibitem[{Julian et~al(2021)Julian, Swanson, Sukhatme, Levine, Finn, and
  Hausman}]{Julian_2021_Never}
Julian R, Swanson B, Sukhatme G, et~al (2021) Never stop learning: The
  effectiveness of fine-tuning in robotic reinforcement learning. In: Kober J,
  Ramos F, Tomlin C (eds) Proceedings of the 2020 Conference on Robot Learning,
  Proceedings of Machine Learning Research, vol 155. PMLR, pp 2120--2136,
  \urlprefix\url{https://proceedings.mlr.press/v155/julian21a.html}

\bibitem[{Jumper et~al(2021)Jumper, Evans, Pritzel, Green, Figurnov,
  Ronneberger, Tunyasuvunakool, Bates, {\v{Z}}ídek, Potapenko
  et~al}]{Jumper_2021_Highly}
Jumper J, Evans R, Pritzel A, et~al (2021) Highly accurate protein structure
  prediction with alphafold. Nature 596(7873):583--589

\bibitem[{Khosla et~al(2011)Khosla, Jayadevaprakash, Yao, and
  Fei-Fei}]{Khosla_2011_Novel}
Khosla A, Jayadevaprakash N, Yao B, et~al (2011) Novel dataset for fine-grained
  image categorization. In: First Workshop on Fine-Grained Visual
  Categorization, IEEE Conference on Computer Vision and Pattern Recognition,
  Colorado Springs, CO,
  \urlprefix\url{http://vision.stanford.edu/aditya86/ImageNetDogs/}

\bibitem[{Kingma(2014)}]{Kingma_2014_Adam}
Kingma D (2014) Adam: a method for stochastic optimization. arXiv preprint
  arXiv:14126980

\bibitem[{Krizhevsky et~al(2009)Krizhevsky, Hinton
  et~al}]{Krizhevsky_2009_Learning}
Krizhevsky A, Hinton G, et~al (2009) Learning multiple layers of features from
  tiny images

\bibitem[{Li et~al(2023)Li, Liu, Sigal, and Liao}]{Li_2023_GraphPNAS}
Li M, Liu JY, Sigal L, et~al (2023) Graph{PNAS}: Learning probabilistic graph
  generators for neural architecture search. Transactions on Machine Learning
  Research \urlprefix\url{https://openreview.net/forum?id=ok18jj7cam}

\bibitem[{Liu et~al(2019)Liu, Simonyan, and Yang}]{Liu_2019_Darts}
Liu H, Simonyan K, Yang Y (2019) {DARTS}: Differentiable architecture search.
  In: International Conference on Learning Representations,
  \urlprefix\url{https://openreview.net/forum?id=S1eYHoC5FX}

\bibitem[{Mills et~al(2023)Mills, Niu, Salameh, Qiu, Han, Liu, Zhang, Lu, and
  Jui}]{Mills_2023_AIOP}
Mills KG, Niu D, Salameh M, et~al (2023) Aio-p: Expanding neural performance
  predictors beyond image classification. Proceedings of the AAAI Conference on
  Artificial Intelligence 37(8):9180--9189. \doi{10.1609/aaai.v37i8.26101},
  \urlprefix\url{https://ojs.aaai.org/index.php/AAAI/article/view/26101}

\bibitem[{Pardo et~al(2018)Pardo, Tavakoli, Levdik, and
  Kormushev}]{Pardo_2018_Time}
Pardo F, Tavakoli A, Levdik V, et~al (2018) Time limits in reinforcement
  learning. In: Dy J, Krause A (eds) Proceedings of the 35th International
  Conference on Machine Learning, Proceedings of Machine Learning Research,
  vol~80. PMLR, pp 4045--4054,
  \urlprefix\url{https://proceedings.mlr.press/v80/pardo18a.html}

\bibitem[{Pham et~al(2018)Pham, Guan, Zoph, Le, and Dean}]{Pham_2018_Efficient}
Pham H, Guan M, Zoph B, et~al (2018) Efficient neural architecture search via
  parameters sharing. In: Proceedings of the 35th International Conference on
  Machine Learning, pp 4095--4104,
  \urlprefix\url{https://proceedings.mlr.press/v80/pham18a.html}

\bibitem[{Power et~al(2021)Power, Burda, Edwards, Babuschkin, and
  Misra}]{Power_2021_Grokking}
Power A, Burda Y, Edwards H, et~al (2021) Grokking: Generalization beyond
  overfit-ting on small algorithmic datasets. In: 1st Mathematical Reasoning in
  General Artificial Intelligence Workshop at the International Conference on
  Learning Representations,
  \urlprefix\url{https://mathai-iclr.github.io/papers/papers/MATHAI\_29\_paper.pdf}

\bibitem[{Qian et~al(2023)Qian, Qin, Luo, Wang, and Wu}]{Qian_2023_Deep}
Qian Q, Qin Y, Luo J, et~al (2023) Deep discriminative transfer learning
  network for cross-machine fault diagnosis. Mechanical Systems and Signal
  Processing 186:109884. \doi{https://doi.org/10.1016/j.ymssp.2022.109884},
  \urlprefix\url{https://www.sciencedirect.com/science/article/pii/S0888327022009529}

\bibitem[{Schaul et~al(2016)Schaul, Quan, Antonoglou
  et~al}]{Schaul_2016_prioritized}
Schaul T, Quan J, Antonoglou I, et~al (2016) Prioritized experience replay
  [c/ol]. In: Proceedings of the 4th Inter national Conference on Learning
  Representations, ICLR

\bibitem[{Tan and Le(2019)}]{Tan_2019_Efficientnet}
Tan M, Le Q (2019) Efficientnet: Rethinking model scaling for convolutional
  neural networks. In: International conference on machine learning, Pmlr, pp
  6105--6114

\bibitem[{Tan et~al(2019)Tan, Chen, Pang, Vasudevan, Sandler, Howard, and
  Le}]{Tan_2019_MnasNet}
Tan M, Chen B, Pang R, et~al (2019) Mnasnet: Platform-aware neural architecture
  search for mobile. In: Proceedings of the IEEE/CVF Conference on Computer
  Vision and Pattern Recognition (CVPR),
  \urlprefix\url{https://openaccess.thecvf.com/content\_CVPR\_2019/html/Tan\_MnasNet\_Platform-Aware\_Neural\_Architecture\_Search\_for\_Mobile\_CVPR\_2019\_paper}

\bibitem[{Tu et~al(2022)Tu, Roberts, Khodak, Shen, Sala, and
  Talwalkar}]{Tu_2022_NASBench360}
Tu R, Roberts N, Khodak M, et~al (2022) Nas-bench-360: Benchmarking neural
  architecture search on diverse tasks. In: Koyejo S, Mohamed S, Agarwal A,
  et~al (eds) Advances in Neural Information Processing Systems, vol~35. Curran
  Associates, Inc., pp 12380--12394,
  \urlprefix\url{https://proceedings.neurips.cc/paper\_files/paper/2022/file/506630e4a43bb9d64a49f98b9ba934e9-Paper-Datasets\_and\_Benchmarks.pdf}

\bibitem[{Vaswani et~al(2017)Vaswani, Shazeer, Parmar, Uszkoreit, Jones, Gomez,
  Kaiser, and Polosukhin}]{Vaswani_2017_Attention}
Vaswani A, Shazeer N, Parmar N, et~al (2017) Attention is all you need. In:
  Guyon I, Luxburg UV, Bengio S, et~al (eds) Advances in Neural Information
  Processing Systems, vol~30. Curran Associates, Inc.,
  \urlprefix\url{https://proceedings.neurips.cc/paper\_files/paper/2017/file/3f5ee243547dee91fbd053c1c4a845aa-Paper.pdf}

\bibitem[{Williams(1992)}]{Williams_1992_Simple}
Williams RJ (1992) Simple statistical gradient-following algorithms for
  connectionist reinforcement learning. Machine learning 8:229--256

\bibitem[{Ying et~al(2019)Ying, Klein, Christiansen, Real, Murphy, and
  Hutter}]{Ying_2019_NASBench101}
Ying C, Klein A, Christiansen E, et~al (2019) {NAS}-bench-101: Towards
  reproducible neural architecture search. In: Chaudhuri K, Salakhutdinov R
  (eds) Proceedings of the 36th International Conference on Machine Learning,
  Proceedings of Machine Learning Research, vol~97. Pmlr, Long Beach,
  California, USA, pp 7105--7114,
  \urlprefix\url{http://proceedings.mlr.press/v97/ying19a.html}

\bibitem[{Zhou et~al(2024)Zhou, Wang, Feng, Liu, Wong, and
  Tan}]{Zhou_2024_Toward}
Zhou X, Wang Z, Feng L, et~al (2024) Toward evolutionary multitask
  convolutional neural architecture search. IEEE Transactions on Evolutionary
  Computation 28(3):682--695. \doi{10.1109/tevc.2023.3348475}

\bibitem[{Zhu et~al(2023)Zhu, Lin, Jain, and Zhou}]{Zhu_2023_Transfer}
Zhu Z, Lin K, Jain AK, et~al (2023) Transfer learning in deep reinforcement
  learning: A survey. IEEE Transactions on Pattern Analysis and Machine
  Intelligence 45(11):13344--13362. \doi{10.1109/TPAMI.2023.3292075}

\bibitem[{Zoph and Le(2017)}]{Zoph_2017_Neural}
Zoph B, Le Q (2017) Neural architecture search with reinforcement learning. In:
  International Conference on Learning Representations,
  \urlprefix\url{https://openreview.net/forum?id=r1Ue8Hcxg}

\end{thebibliography}

\end{document}